# AI Technologies in Language Access: Attitudes Towards AI and the Human Value of Language Access Managers

**Miguel A. Jiménez-Crespo**
Rutgers University/ Dept. of Spanish and Portuguese, 15 Seminary Pl 5th fl., New Brunswick, NJ, 07030, USA
Jimenez.miguel@rutgers.edu

**Stephanie Rodriguez**
Rutgers University/ Dept. of Spanish and Portuguese, 15 Seminary Pl 5th fl., New Brunswick, NJ, 07030, USA
srodrig@newark.rutgers.edu

**Alejandro Jaume**
Rutgers University/ Dept. of Spanish and Portuguese, 15 Seminary Pl 5th Floor fl.USA
ajaumelosa@spanport.rutgers.edu

## Abstract

The rapid emergence of AI technologies is reshaping translation practices and theory across the board. This paper deals with the impact of AI in language access. This area is characterized by the need to serve broad and diverse user populations, within a context where efficiency and access are shaped by legal mandates, ethical and commercial tensions, and safety concerns. This paper reports on the attitudes and perceptions of language access managers towards the AI and the human value in the AI age. Methodologically, this paper presents an analysis of a subset of a broader study on language access and technology, specifically a qualitative thematic analysis of ten semi-structured interviews with language access managers working in healthcare, court, public service and local government contexts. The results indicate that language access managers show conditional optimism towards the inevitable AI implementations, are strongly risk aware, and deeply committed to the human value and human oversight of AI implementations and output.

## 1 Introduction

Language Access is a key priority for governments, organizations and institution across the world. In the USA, recent legal initiatives such as NJ (S2459- 2023) and NY state (Section 202a - 2022) require government agencies, as well as those receiving state funds such as hospitals or non-profits, to translate vital documents into the most spoken languages. In the USA, approximately 22% of people speak a language other than English at home while 8-9% identify as speaking English "less than very well" (25.7 million) (U.S. Census). In the Tri-State area, approximately 30% of residents speak a language other than English at home, while 13% of NY residents and 5.2 % of NJ residents have limited English proficiency (LEP). Nevertheless, little is known to date about how these governmental bodies, institutions and non-profits in the U.S. make use of new language translation technologies for critical and non-critical communicative needs, such as neural machine translation (NMT), generative large language models (LLMs) or automatic speech recognition (ASR). Recent studies show that these technologies help bridge certain immediate communicative gaps that might otherwise remain unresolved, and they are frequently used by both providers and Limited English Proficiency users (LEP) (Bowker 2023: 92). Given the rapid advancements in language technologies, the use of these technologies across different public settings requires immediate attention.

Language access represents a distinct and specific research area given its impact on societies. Meaningful language access requires more than the availability of language professionals or technology integration in workflows. It also depends on institutional commitment that calls for administrative implementation and formal policy frameworks that embed language access within organizational structures (Flores 2005; Angelelli 2015). Historically, such efforts have been shaped by monolingual assumptions that favor dominant-language speakers and sideline speakers of non-dominant languages, reinforcing systemic disparities in access to public services (Mellinger and Monzó-Nebot 2022). Addressing this requires staff and administrators to actively advocate for LEP individuals and highlight the importance of language access in these domains, which will build institutional cultures where language service professionals are recognized as essential to public access and communication, and work towards breaking down language barriers (Wang, 2009).

Providing access to language services for LEP individuals is generally negotiated based on resources framed around the allocation of funding (Mellinger & Monzó-Nebot, 2022). Despite the growing use of translation technologies across many fields, their application in public service settings remains constrained by financial, institutional, and textual limitations (Killman and Mellinger, 2022). These tools also raise concerns around data privacy, biases, rights protection, and translation quality (Monzó-Nebot and Tasa-Fuster, 2024).

Studies have started to research the perceptions and attitudes towards the use of MT or AI of the wider range of stakeholders in the communicative event in public settings, such as healthcare professionals (Gonzalez et. al 2024; Valdes 2025) or end users, including for example migrant communities (Yang and Valdes 2025). The urgency of this research question is due to serious risks involved in technology-mediated communicative events in critical contexts and the need to guarantee language access rights. These risks range from low acceptance rates, misinformation, gender or racial biases, lack of end-user engagement, confidentiality, ethical or language justice issues. These problems, among others, can seriously impact disadvantaged language communities and further exacerbate the language access gap in globalized multilingual societies.

This study reports on the perceptions and attitudes towards AI and translation technologies in language access by one of the key stakeholder groups, language access managers. According to the U.S. Department of Justice (2011), Language Access Managers are professionals responsible for designing, implementing, and overseeing language access policies and programs to ensure that individuals with limited English proficiency (LEP) can meaningfully access services, information, and legal protections provided by public agencies and institutions. This role typically involves conducting language needs assessments, coordinating interpreter and translation services, developing compliance strategies with civil rights legislation (such as Title VI of the Civil Rights Act and Executive Order 13166), training staff on effective multilingual communication, and monitoring language service quality to reduce linguistic barriers in healthcare, legal, and social services settings. Compliance with U.S. federal legislation in terms of language access is required for all those bodies and institutions that receive any type of federal funding.

The study of the professionals who oversee workflows and processes is necessary, given that most recent publications on attitudes towards AI have focused solely on professional translators (Rivas Ginel and Moorkens 2024; Jimenez-Crespo 2024, 2025), while others include language service provides (LSPs) (GALA 2025; ELIS 2025) or students and trainees (ELIS 2025). These studies show that perceptions and use of AI in translation depend on a wide variety of factors. One of these factors being the perceived imposition by those professionals that manage organizations, such as translation, localization, or language access managers, LSPs, digital platform managers, etc. (e.g., Jimenez-Crespo 2025). As such, developing a clearer picture of current perceptions requires an analysis of those responsible for managing and coordinating translation and interpreting work. These professionals work at a unique intersection of language access, safety, liability, budgetary constraints and regulatory mandates.

## 2 The study: research questions and methodology

This study forms part of a wider global project examining translation and language technologies for language access with the objectives of researching current use, attitudes and perceptions, as well as mapping training needs in this area. The present paper focuses specifically on attitudes towards AI within the dataset collected for this broader project, which includes an online survey and semi-structured interviews with language access managers in court, health, and state and local governments.

## 2.1 Research questions

| Theme | Subthemes |
|---|---|
| **Perception of AI (attitudes, framing)** | AI as Risk Object |
| | AI as Inevitable |
| | AI for Productivity |
| | AI as an Assistant/Friend |
| | AI and Human Superiority / Favoritism |
| | AI and human supervision |
| | Ethical Tensions: Access/ safety/liability/ cost |
| | AI as difficulty |
| | AI and quality |
| | AI as industry hype |
| | AI and difference between translation and interpreting |

Table 1: Themes and subthemes related to perception and attitudes towards AI.

The following research questions are addressed in this paper:

- What are the attitudes and perceptions of language access managers towards AI?
- How do language access managers perceive the role of human agents (e.g., professionals, bilingual staff, everyday/untrained?? users, and community members) in the AI era?

## 2.2. Methodology

For the broader multimethod study, an online survey and a semi-structured interview script were developed. Data from both instruments were collected between October and December 2025. The survey was designed using the online platform Qualtrics, while the script for the semi-structured interviews was adapted from the interview questions for the European LT-Lider project (Secară et al. 2025). Ethical approval was obtained from the Institutional Review Board at Rutgers University, with the protocol approved on 8-18-2025. A combination of convenience and snowball sampling (Bernacki and Waldorf 1981) was used. Potential participants were recruited through Google and LinkedIn searches for job profiles such as “language access coordinator”, “language access manager” or related titles. The online survey included a final question asking about their availability to participate in a semi-structured interview. Participants who responded with willingness were subsequently contacted to schedule the interview. Only those professionals who held full-time language access management positions in courts, hospital and school systems, as well as government/ public service were invited. The interviews were conducted via Zoom, and were recorded, transcribed and proofed. All interview data were anonymized to ensure participant privacy. Prior to participation, a brief introduction outlining the purpose of the study, duration, data handling, privacy protocols, and voluntary participation were shared with participants. Informed consent was obtained online before participants proceeded with the interview. The interviews were organized into three thematic sections. The first section focused on technology use and technology adoption. The second addressed automation and quality. The third and last section examined competences and skills. This final section was included given the overall secondary objective of the study, which was to map technology related skills and competences required in the content of the AI era.

The interviews were initially examined by three members of the research team and they were coded using thematic content analysis (Braun and Clark 2006). The coding scheme was developed inductively based on patterns identified in the responses, resulting in an initial set of themes and subthemes based on similar responses across the dataset. After the initial set of themes was discussed and refined in series of research team meetings in which differences in interpretation were addressed and the coding framework was further developed. This process resulted in a final set of themes and subthemes. The authors then categorized all responses using the final revised coding scheme. Any remaining differences in coding were discussed and resolved through consensus to ensure intercoder reliability.

**Themes and subthemes**

As previously noted, the bottom-up inductive analysis of these future-oriented questions was iterated and resulted in a range of themes and subthemes. The following list includes the main themes related to attitudes related to (1) attitudes and perceptions towards AI, and (2) the added value or role of humans in the AI age. Other themes of interest to the intersection of AI and language access are themes related to adoption or language-based differences and limitations, but due to space constraints they will be reported in future publications. The themes (2) and subthemes (19) reported in this paper are listed here in order of frequency.

## 3 Results

The analysis is organized into two sections, which have been previously described: (1) the perceptions of language access managers towards AI, and (2) the perceived human value in the AI age.

## 4 Perception of AI (attitudes, framing)

### 4.1 AI and risk

The main subtheme that dominates the discourse of participants in the study is "AI as risk". Risks associated with AI also frame related issues such as technological adoption, privacy and security concerns, as well as subthemes that emerge from ethical tensions in terms of safety vs access. This finding aligns with previous studies examining the risk of MT use (e.g. Nuurminen and Koponen 2024). The risks discussed by language managers relate to both end users and communities, as well as to organizations themselves. Risks to organizations depend on the setting (legal, government, medical) and the different legal and compliance environments in which AI tools get deployed. For example, in medical settings, Participant 7 (P7) indicates that the potential economic savings from AI deployment might not be justified if the risk involves potential lawsuits to the organization: "If anything goes wrong, we are going to have to pay millions of dollars with a lawsuit". Similarly, P3 discusses how a key driver of AI implementations is an analysis of potential "risk to patient harm if this is used in areas where it should not be". In healthcare contexts, potential risks to end users are therefore considered alongside risks to the organization, whereas concerns about risks to users appear particularly marked in government and legal setting.

An interesting issue is how participants discuss differences in high vs low risk situations (e.g. Guerberof-Arenas and Moorkens 2023). Participants describe how decisions are made in the organization, including workflow charts to differentiate situations such as "first point of contact" (P9) or directing users to bathrooms or service areas (P8) where risk is considered minimal. As P9 explains, AI can be used: "For low-risk situations, just like providing directions, sort of like we did meet and greet with, clients. Some of the language technology would be easily employed just to have a long term, support for those first points of contact" (I9).

| Themes | Subthemes |
|---|---|
| **Human value in the ai age** | Context & Nuance Sensitivity |
| | Tone & Emotional Reading |
| | Empathy & Compassion |
| | Pscychological states |
| | Cultural Brokerage |
| | Sensitive topics/ taboo |
| | Language change |

Table 2: Themes and subthemes related to humans value in the AI age.

A related issue in large organizations are risks related to digital implementation within the broader technological ecosystem of organizations and cybersecurity. This issue is part of a separate research question in the wider study related to adoption of technology. Cybersecurity within organizations, as well as data protection and privacy are the main drivers shaping decisions about the adoption and implementation of technological tools (e.g. Canfora and Ottmann 2020). Often, language managers depend on other units within the organizations to approve the use and integration of any technological tool. Similarly, participants also expressed reservations about working with external vendors, as they might not guarantee data privacy and security, often due to regulation or directives from tech divisions in organizations. For example, P9 expresses how they cannot work with international vendors, despite economic savings, because “they don't meet the [safety] requirements” (P9) required by their IT unit. Similarly, language managers express that even when they provide guidance and risk decision charts they cannot control if all units or employees consistently use secure and privacy compliant AI tools (e.g., MT, LLMs, STT technologies).

## 4.2 AI as Inevitable and industry hype

Despite the risks and challenges identified, participants discuss throughout the interviews that AI is inevitable, or as P1 indicates “It is unstoppable. It is here.” (P1). Organizations and translators, despite conducting risk assessments and operating within organizational limitations, indicate that “we have to find a way to have a relationship with it.” (P1). Participants express that technology is inevitably going to be part of their daily work (P3), and to some extent, they cannot imagine their work otherwise as indicated by P1: "I use it every day. I can't imagine going back." Despite this inevitability, the levels of implementations vary due to a wide range of constraints, such as performance limitations, cybersecurity concerns, budget constraints, safe integration with existing systems, end user or community reception, etc. Some participants, such as P9, see it as a process of exploration today, “we're very aware of a lot of the technologies that are out there and exploring some of what some are possibilities to bring in specifically for our team.” In addition, several participants indicate that they conducted studies and pilots and they failed miserably (P7). This sense of inevitability is often balanced by a perception of industry hype and the promotion of technologies by vendors (P8), as well as by advertising claims made by companies (P2, P3).

Another subtheme that relates to the inevitability of AI is the differences between translating and interpreting. Language managers oversee organizations with different level of needs in terms of what type of mediation needs and most participants mentioned this dichotomy refers to the lack of maturity of interpreting technologies. For example, when asked whether AI could be used for interpreting, P7 responds that, “for translation I do, not for interpreting”, adding that privacy concerns remain critical, particularly regarding “where the information is stored”. Similarly, P1 indicate that for interpreting in courts it is far from achieving real implementations, while “for translations… I cannot live without it now”.

This theme is counterbalanced by another theme related to industry hype, previously mentioned. Such hype is a widespread phenomenon and is related to the push for investment in the AI technology sector (Hanna and Bender 2025). Participants directly addressed this hype, often related to failed pilot project that had been promoted by language technology vendors. For example, P9 indicates “there are always limitations and it's never quite as seamless as […] the marketing or the hype.” Similarly, P3 refers to their marketing efforts “There’s just so much marketing… you can just put the earbuds in, and it does all the work.” The hype also contributes to anxieties about the future of the profession and the perception that everyone is somewhat fearful of AI, or in the words of P7 “like AI is the end of the world”.

## 4.3 AI for productivity and AI as an assistant

Two themes that are interrelated are the perception of AI as a tool for productivity and the use of AI as an assistant or friend of the human. The role of AI within the wider language industry ecosystem has been reported by GALA Automation Barometer (2025). Language service providers are increasingly leveraging AI for several management tasks other than translation proper, which appears currently as the most impactful implementation of AI in the industry. Similarly, survey studies with translation professionals have identified as the most frequent uses of AI augmentation (Jiménez-Crespo 2025) or AI-assisted brainstorming and inspiration (Rivas Ginel and Moorkens 2024). This is reflected in language access managers attitudes, as participants express it truly helps to “streamline the processes” (P3) and make their work “more efficient” (P1). Efficiency and higher quality are often mentioned as the benefits of higher productivity, as indicated by P9: “[linguists are] actively exploring different tools and how to, how to help them improve their work, either […] to make it faster or easier to deliver quality, translations.”

This is directly related to the theme that conceptualizes AI as an assistant or a friend, in terms of augmenting human performance to "expand our reach and do more" (P3). Some participants directly use the term "friend" in their responses, such as P3: "AI can really be our friend if we know how to use it". As previously mentioned, productivity relates to any and all components of the language access program, from management to preparation of glossaries or assets all the way to the actual translation process (GALA 2025). In terms of overall project management, P3 expresses how AI can be "a big-time savings for the project management piece". In their responses, language access managers refer both to managing translation and interpreting tasks, and in both cases the use of AI is perceived as a tool to manage and facilitate both human and digital resources.

### 4.4 AI, human superiority and human supervision

The perceived superiority in terms of performance and decision making of humans, which is understood here in broad terms as professional, non-professionals and community members, emerged as a key theme in the interviews. This ties on the one hand with results of a myriad of studies in translation studies (e.g. Guerberof Arenas and Toral 2022; Kenny and Winters 2024), as well as with the need to counter the human parity claims in the industry (Toral 2019; Yan et al 2024). Overall, participants expressed the preference for human translators and interpreters despite resource and economic constraints: "We would rather use humans". This is especially important in interpreting tasks, where speech to speech (STS) technologies are perceived as not ready for language access implementations in high-risk settings (e.g. Elsayed 2026). As P1 indicates, "There is nothing better than in person interpreters".

The perception that human performance is perceived as superior to current AI solutions means that for many use cases AI is not perceived as a replacement, even when this is a common fear in the profession (Jiménez-Crespo 2025), amplified by different widely distributed industry reports (Tomlison et al 2025). In real-life language access scenarios, participants indicate that they perceive AI as "a tool, but it's not a replacement." (P6). The use of AI as a tool, together with the superiority of human performance leads to the related theme of human supervision.

In this next related theme, AI is perceived as an object of human supervision, often related to post editing or revision paradigms. Humans are perceived as essential in the process because there is "always a need for a qualified human reviewer" (P6). The process often involves post editing, even when sometimes established practices such as backtranslation in medical settings is still in use (Bundgaard and Brogger 2018). For example, P7 indicated that for critical materials, this is still the process they follow to assess the adequacy of AI-driven translations, even when edited. Participants emphasize a perceived need "to understand what the risks are if we don't edit" (P4), because otherwise a large number of errors can be left in the final product. This means that there can be serious consequences, such as compromising "patient safety".

### 4.5 Ethical tensions related to access vs safety, liability or cost

Ethical issues are a key issue in research on AI implementation in general (Capel and Brereton 2023), as well as in translation (e.g. Moorkens 2025). In extant research, ethical issues are often related to gender, racial or Western biases, as well as data privacy or date related issues. In this study, the practical approach of language access managers means that the ethical tensions are primarily related to how AI is implemented in their organizations, such as the balancing access to services and safety. As P4 indicates, the market is "starving for instant translation", but asks about that "balance of safety or access" because "here's some where there's a very thin line, and the ethicists are the ones that need to come and look at that." This balance is generally addressed in terms of potential risks to both end users and organizations. Normally, the possibility of expanding access to services using technologies is perceived as a positive, but often "the risks are too great" (P4). This tension for managers is shaped directly also by limitations such as cost or liability issues. For some organizations, cost savings imply that AI tools will certainly be used "If we can save $2 million with this new application… we are going to make it work." (P7). Nevertheless, this same participant also argues in the opposite direction. P7 also indicate that organizational liability can stop

this due to potential lawsuits, “You don’t want to save a few thousand dollars and then pay a million later.”

These tensions shape the adoption and use of AI across language access context. Decisions about adoption is a responsibility shared with a number of teams within organizations, where safety or access can be determined by legal, business or technological units. For example, while AI tools might increase efficiency in language access programs, IT departments may prohibit their use due to cybersecurity risks. Such risks may arise when the servers of products and apps might not be secure or in the right geographical location (P9). Similarly, the potential expansion of access through outsourcing translation tasks to other countries might be stalled if the organization forbids the use of international vendors due to these same risks. Nevertheless, it is also acknowledged that working with vendors sometimes the safety and privacy risks cannot be fully controlled as outside vendors might not disclose the use of LLM or MT technologies (P4). Consequently, this ethical tension between access, safety, cost and liability is seen by one participant as one of the reasons why organizations have “been so slow to respond to the rapid changes in the industry” (P9).

### 4.6 AI attitudes: a summary

The attitudes and perceptions of language access managers can be summarized within a paradigm in which risks dominate the discourse, especially in healthcare settings. Across all interviews the superiority of human translation and interpreting is reaffirmed, but AI is perceived as a tool for productivity and to enhance efficiency in certain processes, but not all. This view of AI as a friend or assistant is also paired with deep skepticism in some cases, especially in interpreting scenarios. As such, the attitudes towards AI in language access environments show that its integration in translation is more acceptable that in interpreting.

## 5 Human value in the AI age

One of the main areas of interest within translation studies is the human value with successive waves of technological innovation. When neural machine translation (NMT) improved dramatically the fluency of output, scholars insisted on a research agenda that highlighted the significance of human agents in the translation ecosystem, or what some have called “the added value of human translation” (Massey & Ehrensberger-Dow, 2017, p. 308). Today, the rise of GenAI has also led to calls to “rethink critically the value and values that humans bring to the translation process.” (Pennet, Moorkens and Yamada 2026: iv). The value of humans in language access is consistently described by participants not in abstract terms, but in highly embodied, communicative and relational terms. They frame this value as a mix of contextual intelligence, an awareness of the relational and personal dimensions of translation and interpreting, as well as the ability to make ethical decisions. These are the main issues that drive participant resistance to full implementation of AI automation.

### 5.1 Context and nuance

The relation of text and context in the communicative setting is one of the key issues in MT, and to a lesser degree, LLMs. Participants describe the human value primarily as the ability to detect nuances and to relate the message to the context. As a whole, participants indicate that AI is “not very good at looking at the whole picture” (P8) and it “does not have an understanding of context” (P8). As a result, issues arise in both in comprehension and in reformulation. Regarding comprehension, it is indicated that humans are better at “[u]nderstanding what the other person means in their context” (P2). During interpreting there are many nuances that “the human could detect”, because of its better understanding of context (P1). In terms of reformation, participants argue that the specifics of the target audience are not fully captured by the linguistic and cultural context of the communities that each language manger needs to address. P8, for example, indicates that the tools are “great […] but need to be shaped for our context". Similarly, P9 discusses the impact on reception of these tools and the need to understand how targeted communities perceive and receive these messages. This observation is directly related to a subtheme within the theme of adoption of technologies related to standard varieties and community specifications. One key issue is that tools are normally trained with standard language varieties, but language and cultural communities in some areas and cities are predominantly speakers of a language variety that does not match what regular tools produce.

Participants identified this challenge particularly in relation to Arabic and several African languages (P8). The need to have this nuance and contextual knowledge to "target to our audience" (P9) is related to the potential lack of trust of the community in government. In fact, the adaptation to the linguistic, cultural and contextual features of the local community is related to whether that community will trust the organization or government behind it. As P8 claims, this contextual knowledge and adaptation is key as "we need people to trust us".

Two related subthemes here are cultural brokerage and language change/ dialects. As far as the first one, the knowledge of the communicative context includes the subtheme cultural brokerage, as humans are "able to identify… a cultural misunderstanding." (P6). They also fully understand cross cultural communication and can identify "what words resonate" in a specific community "how to localize them" (P6). This cultural localization is also seen as something that needs to be community specific, a very specific targeted population (P9). This localization, like what happens with in country reviews of translated content, needs to be tailored to the specific community. For example, P9 discussed how in "the traditional LSP is setting […], linguists overseas" might be "cheaper […] but they are not familiar with how things are... about [how] languages used locally" (I9).

In this deep knowledge of the local community and their context language change and dialectal variation plays a key role. Participants indicate that humans are better than machines because these "[l]ack the nuance of the dialect or variation" (P8). The issue of data for training is highlighted, because oftentimes the output of automatic translation systems might be wrong, but "they're so prevalent everywhere that now have become standard." (P1). To solve this, P8 and P9, both in local government, highlight the necessary involvement of the community to help improve the technology. P9 for example indicates that they are trying to "get community feedback and figure out how best to integrate that into, the tools", and for that they need a dynamic system "where you can update terminology or which will change over time".

### 5.2 Empathy, compassion and human connection

Human empathy and compassion emerge as key components of language access brokering. Participants indicate that in this specific context, humans are needed because they understand "the human condition the way a machine never can" (P2). In many cases compassion is required because oftentimes professionals "[are] dealing with very difficult topics" (P2). These difficult topics also leads to language mediation with participants that are "kind of sputtering and suffering with anxiety" (P2), and in these instances compassion is key. For those cases, AI is perceived as "divorced from the human context" (P2), while with a human the translator and interpreter is always there. The human not only empathizes with the community members, but she can also better understand them because they "can read body expressions, facial expressions … a machine doesn't." (P7). In addition, in interpreting the tone and emotional reading changes the meaning of utterances, and the interpreter in legal settings "have to convey that" (P1). Compassion is also extended to members of the community with low literacy who also require assistance with the use of technologies. For example, some segments of the communities that language access serves might be in very difficult circumstances, and in some cases the use of technology might make communication more difficult if there is no way to solve technological issues or misunderstandings. Overall, this can be summarized as P2 indicates that "what we call that warm, fuzzy human factor".

This human connection thus, extends to detecting different psychological or mental states and how they impact communication. For example, P1 indicated that humans can detect and adapt to "Somebody with declining cognitive function or schizophrenic with hallucinations". In these cases, speakers might not "speaking correctly and getting their words out or like all of those other things" and understanding and dealing with these issues "come in with human experience" (P4).

A related subtheme to this human empathy and compassion relates to how to deal in specific cases with sensitive topics or taboo. P1 indicates that this is quite difficult "especially like sex cases… there are so many euphemisms." Adapting to the specifics of a local community is also difficult when dealing with sensitive topics or those even for which the local community might not have the terminological or conceptual tools to describe or discuss certain topics (P8). In such cases, P8 indicated that work with local communities was key to create glossaries with inclusive, gender-neutral language or LGTBQ terminology. In addition, P2 discussed guardrails in LLMs and the potential to disrupt difficult conversations in legal or medical settings if LLMs are not adapted or trained for these specific contexts. For example, P2 claims that in an AI could be:

> "telling you to stop thinking about violent things and what if it and so like if you could imagine if you're using it for interpreting, it here's something. And instead of transcribing the message it's putting out the information for like a suicide hotline or domestic violence, right?" (P2).

Thus, human value in this case relates to both the adaptation to the local community of how certain topics or language is used, as well as how to deal with LLM guardrails that might render their implementation useless in critical scenarios in medical or legal settings.

## 6 Conclusions

This study has reported the results of qualitative study based on a thematic analysis of semi-structured interviews with ten language access managers in the United States. It is part of a wider study on language and translation technologies in language access, which examines current use, attitudes, and training mapping needs in this area. The study has identified that language access managers perceive AI as inevitable and they are not opposed to implementations and integrations, but they are divided between optimism and caution, often depending on the sector and whether the organization provides more translation or interpreting. They are consistently strongly risk aware, and thus they are deeply committed to workflows that include human oversight. This includes not only professionals but also community members that often possess the cultural, linguistic and contextual knowledge necessary to adapt messages to specific linguacultural communities.

In terms of human value, managers consistently perceive that humans are needed for nuance and contextual understanding, including cultural, linguistic or how to handle and navigate sensitive topics or language change. Similarly, they insist on the embodiment of translation and interpreting tasks, with humans showing the necessary empathy, compassion and perception of mental states necessary to mediate in highly stressful and difficult situations for those in need.

The main limitations of the present study relate, first, to the relatively small number of participants, and second, to the selective presentation of only two of the seven themes identified in the broader dataset. The limited participant pool is partly explained by the relatively recent emergence of language access management as a professional role within public service settings in the United States. The selective focus on two themes is also a consequence of the scope and aims of the present contribution. A more comprehensive examination of the remaining themes, including technology adoption, language-based limitations, and current patterns of technology use, would provide a broader understanding of the complexities surrounding AI implementation in language access contexts. In addition, future publications will present the results of the online survey component of the broader project, which included both qualitative and quantitative data. These aspects will be addressed in subsequent publications derived from the wider project. Future research could also replicate this study across different geographical regions or extend the analysis to for-profit industry settings in order to compare institutional priorities, perceptions of risk, and approaches to AI integration across sectors.

## Acknowledgments

This study has been funded by a Rutgers Global Grant at Rutgers University.

## References

Angelelli, C. V. 2015. Justice for all? Issues faced by linguistic minorities and border patrol agents during interpreted arraignment interviews. *MonTi: Monografías de Traducción e Interpretación*, 7:181–205.

Angelelli, C. 2019. *Healthcare interpreting explained*. Routledge.

Bender, E. M. and Hanna, A. 2025. *The AI con: How to fight big tech's hype and create the future we want*. Random House.

Biernacki, P. and Waldorf, D. 1981. Snowball sampling—Problems and techniques of chain referral sampling. *Sociological Methods & Research*, 10(2):141–163.

Bowker, L. 2023. *De-mystifying translation: Introducing translation to non-translators*. Routledge.

Braun, V. and Clarke, V. 2006. Using thematic analysis in psychology. *Qualitative Research in Psychology*, 3(2):77–101.

Bundgaard, K. and Brøgger, M. N. 2019. Who is the back translator? An integrative literature review of back translator descriptions in cross-cultural adaptation of research instruments. *Perspectives*, 27(6):833–845.

Canfora, C. and Ottmann, A. 2020. Risks in neural machine translation. *Translation Spaces*, 9(1):58–77. https://doi.org/10.1075/ts.00021.can

Capel, T. and Brereton, M. 2023. Mapping human-centered artificial intelligence research. In *Proceedings of the 2023 CHI Conference on Human Factors in Computing Systems (CHI '23)*, pages 1–23. https://doi.org/10.1145/3544548

ELIS. 2025. *The European language industry survey – ELIS 2025*. http://elis-survey.org/wp-content/uploads/2025/03/ELIS-2025_Results.pdf

Elsayed, N. 2026. Unseen risks of clinical speech-to-text systems: Transparency, privacy, and reliability challenges in AI-driven documentation. *arXiv preprint arXiv:2601.00382.*

Flores, G. 2005. The impact of medical interpreter services on the quality of health care: A systematic review. *Medical Care Research and Review*, 62(3):255–299.

GALA. 2025. *GALA business barometer report: Technology, automation, and AI 2025*. Globalization and Localization Association.

Gonzalez, C., Graves, J. M., Ramos, J., Vavilala, M. S., and Moore, M. 2024. Language access research for community health: Provider perspectives on language access techniques and the role of communication technology. *Journal of Communication in Healthcare*, 17(1):7–14.

Guerberof-Arenas, A. and Toral, A. 2022. Creativity in translation: Machine translation as a constraint for literary texts. *Translation Spaces*, 11(2):184–212.

Guerberof-Arenas, A. and Moorkens, J. 2023. Ethics and machine translation: The end user perspective. In H. Moniz and C. Parra Escartin, editors, *Towards responsible machine translation: Ethical and legal considerations in machine translation*, pages 113–133. Springer.

Jiménez-Crespo, M. A. 2024. Exploring professional translators' attitudes toward control and autonomy in the human-centred AI era. *Revista Tradumàtica*, 22:276–301.

Jiménez-Crespo, M. A. 2025. Human-centered AI and the future of translation technologies. *Information*, 16(5):387. https://doi.org/10.3390/info16050387

Kenny, D. and Winters, M. 2024. Customization, personalization, and style in literary machine translation. In M. Winters, S. Deane-Cox, and U. Böser, editors, *Translation, interpreting and technological change: Innovations in research, practice and training*, pages 59–79. Bloomsbury Academic.

Killman, J. and Mellinger, C. D. 2022. Technologized legal translation and interpreting: Resource potential, availability, and applications. *Revista de Llengua i Dret*, pages 1–8.

Massey, G. and Ehrensberger-Dow, M. 2017. Machine learning: Implications for translator education. *Lebende Sprachen*, 62(2):300–312.

Monzó-Nebot, E. and Mellinger, C. D. 2022. Language policies for social justice—Translation, interpreting, and access. *Just. Journal of Language Rights & Minorities / Revista de Drets Lingüístics i Minories*, 1(1–2):15–35.

Monzó-Nebot, E. and Tasa-Fuster, V., editors. 2024. *Gendered technology in translation and interpreting: Centering rights in the development of language technology*. Taylor & Francis.

Moorkens, J. 2025. Translation industry ethics. In C. Walker and J. Lambert, editors, *The Routledge handbook of the translation industry*, pages 429–443. Routledge.

Nurminen, M. and Koponen, M. 2020. Machine translation and fair access to information. In *Translation, interpreting and technological change: Innovations in research, practice and training*, pages 111–135. Bloomsbury.

Penet, J. C., Moorkens, J., and Yamada, M., editors. 2026. Introduction. In *Teaching translation in the age of generative AI: New paradigm, new learning?*, pages iii–ix. Language Science Press. https://doi.org/10.5281/zenodo.17580856

Rivas Ginel, M. T. and Moorkens, J. 2024. A year of ChatGPT: Translators' attitudes and degree of adoption. *Tradumàtica*, 22:258–275.

Secară, A., Ginel, M. T., Toral, A., Guerberof, A., Ciobanu, D. I., Brockmann, J., and Rossi, C. 2025. LT-LiDER language technology map: Technologies in translation practice and their impact on the skills needed. https://doi.org/10.25365/phaidra.641

Tomlinson, K., Jaffe, S., Wang, W., Counts, S., and Suri, S. 2025. Working with AI: Measuring the applicability of generative AI to occupations. *arXiv preprint arXiv:2507.07935*.

Toral, A. 2020. Reassessing claims of human parity and super-human performance in machine translation at WMT 2019. In *Proceedings of the 22nd Annual Conference of the European Association for Machine Translation*, pages 185–194.

U.S. Department of Justice. 2011. *Language access assessment and planning tool for federally conducted and federally assisted programs*.

Valdez, S., van Heeswijk, F., and Warren, N. 2025. Machine translation at the hospital: Healthcare professionals' perspectives on use, appropriateness, and policy. *Tradumàtica: Tecnologies de la Traducció*, 23:244–265.

Wang, T. 2009. *Eliminating language barriers for LEP individuals*. Grantmakers Concerned with Immigrants and Refugees.

Yang, T. and Valdez, S. 2025. How machine translation is used in healthcare: Insights from recent Chinese migrants in the Netherlands. *Digital Translation*, 12(2):125–149.